\documentclass[journal,twoside]{IEEEtran}

\usepackage{graphicx}
\usepackage{amsmath}
\usepackage{algorithmic}
\usepackage{url}
\usepackage{bm}

\usepackage{amssymb}
\usepackage{epsfig}
\usepackage{mathtools}
\mathtoolsset{showonlyrefs,showmanualtags}

\makeatletter
\let\NAT@parse\undefined
\makeatother

\usepackage[numbers]{natbib}

\newcommand{\tr}{\mathrm{tr}\,}
\newcommand{\diag}{\mathrm{diag}\,}
\newcommand{\var}{\mathrm{Var}\,}
\newcommand{\cov}{\mathrm{Cov}\,}

\newcommand{\Real}{\operatorname{Re}}
\newcommand{\Imag}{\operatorname{Im}}

\renewcommand{\Psi}{\varPsi}
\renewcommand{\Gamma}{\varGamma}
\renewcommand{\Lambda}{\varLambda}
\renewcommand{\Phi}{\varPhi}
\renewcommand{\Omega}{\varOmega}
\renewcommand{\Sigma}{\varSigma}
\renewcommand{\Theta}{\varTheta}
\renewcommand{\Pi}{\varPi}
\renewcommand{\Upsilon}{\varUpsilon}

\newtheorem{remm}{Remark}

\newtheorem{thm}{Theorem}

\begin{document}

\title{A Note on the SPICE Method}

\author{Cristian~R.~Rojas,~
        Dimitrios Katselis,~%
        and~H\aa kan Hjalmarsson,~\IEEEmembership{Member,~IEEE}
\thanks{C. R. Rojas, D. Katselis and H. Hjalmarsson are with the Automatic Control Lab and ACCESS Linnaeus Center, %
Electrical Engineering, KTH -- Royal Institute of Technology, %
S-100 44 Stockholm, Sweden. %
Emails: {\tt \{cristian.rojas|dimitrios.katselis|hakan.hjalmarsson\} @ee.kth.se}, %
Post: KTH School of Electrical Engineering, Automatic Control, SE-100 44 Stockholm, Sweden.}%
}


\maketitle

\begin{abstract}
In this article, we analyze the SPICE method developed in
\citep{Stoica-Babu-Li-11}, and establish its connections with
other standard sparse estimation methods such as the Lasso and the
LAD-Lasso. This result positions SPICE as a computationally efficient
technique for the calculation of Lasso-type estimators. Conversely,
this connection is very useful for establishing the asymptotic properties
of SPICE under several problem scenarios and for suggesting suitable
modifications in cases where the naive version of SPICE would not work.
\end{abstract}


\IEEEpeerreviewmaketitle

\section{Introduction}
\IEEEPARstart{S}{pectral} line estimation, or the problem of
estimating the amplitudes and frequencies of a signal composed of
a sum of sinusoids contaminated by Gaussian white noise, is a
ubiquitous and well studied area in the field of signal
processing~\citep{Stoica-Moses-05}. Many classes of methods have
been devised to solve this problem under several different
scenarios like, e.g., uniformly/non-uniformly spaced samples,
\emph{a priori} known/unknown number of sinusoids,
homoscedastic/heteroscedastic (constant/varying variance) samples,
parametric/non-parametric model-based, and so on \citep{Stoica-Moses-05,Porat-94,Wang-Li-Stoica-05}.

Recently, SPICE (SemiParametric/SParse Iterative Covariance-based
Estimator), a new technique for spectral line estimation inspired by
ideas from sparse estimation, has been proposed in \citep{Stoica-Babu-Li-11}.
This method is capable of handling irregularly sampled data. Similarly,
a version of SPICE has also been developed for array signal
processing~\citep{Stoica-Babu-Li-11b}, a mathematically almost
equivalent problem~\citep[Chapter~6]{Stoica-Moses-05}.

In this paper, we establish the connection between SPICE
and standard sparse estimation methods such as the
Lasso~\citep{Tibshirani-96} and the
LAD-Lasso~\citep{Wang-Li-Jiang-07}. This connection, based on the
so-called Elfving theorem from optimal experiment
design~\citep{Elfving-52}, puts the SPICE method into perspective,
allowing us to examine the asymptotic properties of SPICE under
several scenarios by simply applying the existing theory for the
Lasso and its variants (see, e.g., the recent
book~\citep{Buhlmann-vandeGeer-11}). Conversely, the relationship
between SPICE and Lasso-type estimators suggests that SPICE may be
used as a (new) numerically efficient technique for computing
Lasso estimates.


The manuscript is organized as follows.
Section~\ref{sec:problem} describes the spectral line estimation
problem and the SPICE method. Section~\ref{sec:analysis}
establishes the relation between SPICE and Lasso-type sparse
estimation methods. In Section~\ref{sec:example} a simulation example
illustrating the equivalence between SPICE and a version of
Lasso is presented. Finally, Section~\ref{sec:conclusions} concludes the
paper.

\emph{Notation:} Vectors and matrices are written in bold lowercase
and uppercase fonts, respectively. $^T$ and $^H$ denote transposition
and complex conjugate transposition, respectively. $\Real z$ and $\Imag z$
stand for the real and imaginary parts of the complex number $z$, and
$j$ is the square root of $-1$. $\mathbb{R}_0^+$ is the set of non-negative
real numbers, and $\mathbb{C}$ is the complex plane.
$\| \cdot \|_1$, $\| \cdot \|_2$, $\| \cdot \|_F$ and $| \cdot |$ correspond to the
$1$-norm, Euclidean norm, Frobenius norm and absolute value, respectively.
$\diag(a_1, \dots, a_n)$ is a diagonal matrix whose diagonal is given by
$a_1, \dots, a_n$. $\bm{I}$ is the identity matrix.
$E\{ \cdot \}$ denotes mathematical expectation.

\section{Problem Formulation and SPICE method} \label{sec:problem}
Consider the following problem: Let $\bm{y} \in \mathbb{C}^{N
\times 1}$ be given, satisfying the equation
\begin{align} \label{eq:model}
\bm{y} = \sum_{k = 1}^K \bm{a}_k s_k + \bm{\epsilon}, %
\end{align}
where $\bm{\epsilon} \in \mathbb{C}^{N \times 1}$ is a complex
Gaussian random vector of zero mean and covariance matrix
$\diag(\sigma_1, \dots, \sigma_N)$, and $\{\bm{a}_k\}_{k = 1}^K
\in \mathbb{C}^{N \times 1}$ are known complex vectors.
$\{s_k\}_{k = 1}^K \in \mathbb{C}$ are unknown complex quantities,
of the form $s_k = |s_k| e^{j \phi_k}$, where the phases
$\{\phi_k\}_{k = 1}^K \in [0, 2 \pi)$ are independent random
variables uniformly distributed in $[0, 2 \pi)$, and the
magnitudes $\{|s_k|\}_{k = 1}^K \in \mathbb{R}_0^+$ are
deterministic parameters to be estimated. The spectral line
estimation problem considers a particular case of
\eqref{eq:model}, where the $\bm{a}_k$'s are vectors of imaginary
exponentials of the form $e^{j \omega t}$~\citep{Stoica-Moses-05}.

In order to estimate the magnitudes $|s_k|$, let
\begin{align} \label{eq:1} %
\bm{R} := E\{\bm{y} \bm{y}^H\} = \bm{A}^H \bm{P} \bm{A}, %
\end{align}
where
\begin{align*}
\bm{A}^H &:= [\bm{a}_1 \; \cdots \; \bm{a}_K \; \bm{I}] \\ %
             &=: [\bm{a}_1 \; \cdots \; \bm{a}_{K + N}] \\ %
\bm{P}   &:= \diag(|s_1|^2, \dots, |s_K|^2, \sigma_1, \dots, \sigma_N) \\ %
             &=: \diag(p_1, \dots, p_{K + N}). %
\end{align*}
The SPICE estimate~\citep{Stoica-Babu-Li-11} of the $|s_k|$'s is
an iterative procedure of the form:
\begin{align} \label{eq:SPICE}
\bm{R}(i) &= \bm{A}^H \diag(p_1(i), \dots, p_{K+N}(i)) \bm{A} \nonumber \\ %
p_k(i + 1)    &= p_k(i) \frac{|\bm{a}_k^H \bm{R}^{-1}(i) \bm{y}|}{w_k^{1/2} \rho(i)}, \qquad w_k := \frac{\|\bm{a}_k\|_2^2}{\|\bm{y}\|_2^2}, \\ %
\rho(i)       &= \sum_{l=1}^{K + N} w_l^{1/2} p_l(i) |\bm{a}_l^H \bm{R}^{-1}(i) \bm{y}|, \nonumber %
\end{align}
where $i$ is the iteration number, and $p_k(i)$ is the estimate of
$p_k$ at iteration $i$. This method is initialized by any initial
estimate of the $p_k$'s, and its estimate $\bm{R}(i)$ converges to
the matrix $\bm{R}$ minimizing
\begin{align} \label{eq:2} %
f(\bm{R}) := \| \bm{R}^{-1/2} (\bm{y} \bm{y}^H - \bm{R})\|_F^2. %
\end{align}
The $p_k$'s that give $\bm{R}$ correspond to the limits
$\lim_{i \to \infty} p_k(i)$.

\begin{remm}
The presence of the inverse of $\bm{R}(i)$ in the SPICE method may
in principle lead to complications if such a matrix becomes singular.
However, if the $p_k(0)$'s are chosen to be strictly positive, then
$\bm{R}(i+1)$ is generically non-singular (since $\bm{a_k}$ is
generically in the column range of $\bm{R}(i)$, and $\bm{y}$ is a
Gaussian random vector which lies in the null space of $\bm{R}(i)$
with probability $0$). Because of this, here and in the sequel we
will implicitly assume for the derivations that $\bm{R}$ is non-singular.
\end{remm}

\begin{remm}
In \citep{Stoica-Babu-Li-11b}, SPICE was defined based
on a slightly different $f(\bm{R})$. We will not consider that
version of SPICE, because such a version can only be defined in a
multi-snapshot case. However, similar steps as the ones described
in the following sections can be applied to the method in
\citep{Stoica-Babu-Li-11b} to arrive at an equivalent Lasso-type
formulation.
\end{remm}

\section{Analysis of SPICE} \label{sec:analysis}
The first version of SPICE in \citep{Stoica-Babu-Li-11} allows the
variances $\sigma_k$ to be different, while a variant of the
method imposes the constraint that $\sigma_1 = \cdots \sigma_N =:
\sigma$~\citep[Section~III.D]{Stoica-Babu-Li-11}. We will treat
these cases separately, starting with the case where the variances
can be different.

\subsection{Different variances} \label{subsec:different_sigma}
As shown in \citep{Stoica-Babu-Li-11}, the function $f$  in
\eqref{eq:2} can be written as
\begin{align*}
f(\bm{R}) &= \tr\{ [\bm{R}^{-1/2} (\bm{y} \bm{y}^H - \bm{R})]^H \bm{R}^{-1/2} (\bm{y} \bm{y}^H - \bm{R}) \} \\ %
              &= \|\bm{y}\|_2^2 \bm{y}^H \bm{R}^{-1} \bm{y} - 2\|\bm{y}\|_2^2 + \tr \bm{R}, %
\end{align*}
hence minimizing $f(\bm{R})$ is equivalent to minimizing
\begin{align} \label{eq:g(R)}
g(\bm{R}) %
&:= \bm{y}^H \bm{R}^{-1} \bm{y} + \frac{1}{\|\bm{y}\|_2^2} \tr \bm{R} \\ %
&= \bm{y}^H \bm{R}^{-1} \bm{y} + \sum\limits_{k = 1}^{K + N} \frac{\|\bm{a}_k\|_2^2}{\|\bm{y}\|_2^2} p_k \nonumber \\ %
&= \bm{y}^H \bm{R}^{-1} \bm{y} + \sum\limits_{k = 1}^{K + N} w_k p_k, \nonumber %
\end{align}
subject to $p_k \geq 0$, where
\begin{align*}
w_k := \frac{\|\bm{a}_k\|_2^2}{\|\bm{y}\|_2^2}. %
\end{align*}
To further simplify the problem, in \citep[Appendix
B]{Stoica-Babu-Li-11b} it is argued that the minimization of
$g(\bm{R})$ is equivalent to solving
\begin{align} \label{eq:3} %
\begin{array}{cl}
\min\limits_{p_1, \dots, p_{K + N} \geq 0} & \bm{y}^H \bm{R}^{-1} \bm{y} \\ %
\text{s.t.}                                & \sum\limits_{k = 1}^{K + N} w_k p_k = 1 \\ %
                                           & \sum\limits_{k = 1}^{K + N} \bm{a}_k \bm{a}_k^H p_k = \bm{R}. %
\end{array}
\end{align}
Equation~\eqref{eq:3} will be our starting point for the analysis
of SPICE. A slight simplification can be achieved by defining
$\tilde{p}_k := w_k p_k$ and $\bm{\tilde{a}}_k := w_k^{-1/2}
\bm{a}_k$ for all $k = 1, \dots, K + N$. This gives the re-parameterized problem
\begin{align} \label{eq:4} %
\begin{array}{cl}
\min\limits_{\tilde{p}_1, \dots, \tilde{p}_{K + N} \geq 0} & \bm{y}^H \bm{R}^{-1} \bm{y} \\ %
\text{s.t.}                                                & \sum\limits_{k = 1}^{K + N} \tilde{p}_k = 1 \\ %
                                                           & \sum\limits_{k = 1}^{K + N} \bm{\tilde{a}}_k \bm{\tilde{a}}_k^H \tilde{p}_k = \bm{R}. %
\end{array}
\end{align}
The strategy now is to consider a derivation similar to Elfving's
theorem, from optimal experiment design~\citep{Elfving-52}, to
obtain an optimization problem equivalent to \eqref{eq:4}. First
notice that
\begin{multline} \label{eq:5} %
\left. \left(\bm{y}^H \bm{R}^{-1} \bm{y}\right) \right|_{\bm{R} = \sum_{k = 1}^{K + N} \bm{\tilde{a}}_k \bm{\tilde{a}}_k^H \tilde{p}_k} \\ %
= \min\limits_{c_1, \dots, c_{K + N}} \sideset{}{'}\sum\limits_{k = 1}^{K + N} \frac{{|c_k|}^2}{\tilde{p}_k} \quad \text{s.t.} \quad \bm{\tilde{A}}^H \bm{c} = \bm{y}, %
\end{multline}
where $\bm{\tilde{A}}^H := [\bm{\tilde{a}}_1 \; \cdots \;
\bm{\tilde{a}}_{K + N}]$ and $\bm{c} := [c_1 \; \cdots \; c_{K +
N}]^T$. Here the $'$ symbol in the summation sign indicates that
the values of $k$ for which $\tilde{p}_k = 0$ should be omitted
from the sum. The proof of \eqref{eq:5} is given in the appendix.

The combination of \eqref{eq:4} and \eqref{eq:5} gives a
minimization problem in $\{\tilde{p}_k\}$ and $\{c_k\}$, i.e.,
\begin{align} \label{eq:7}
\begin{array}{cl}
\min\limits_{\begin{array}{c} \tilde{p}_1, \dots, \tilde{p}_{K + N} \geq 0, \\ c_1, \dots, c_{K + N} \end{array}} %
& \displaystyle \sideset{}{'}\sum\limits_{k = 1}^{K + N} \frac{\displaystyle |c_k|^2}{\displaystyle \tilde{p}_k} \\ %
\text{s.t.}                           & \sum\limits_{k = 1}^{K + N} \tilde{p}_k = 1 \\ %
                                      & \bm{\tilde{A}}^H \bm{c} = \bm{y}, %
\end{array}
\end{align}
where the order of the minimizing variables can be exchanged.
Now, when the $c_k$'s are kept fixed, the minimization of the cost in
\eqref{eq:7} with respect to $\{\tilde{p}_k\}$ can be done
explicitly. To see this, notice that by the Cauchy-Schwarz inequality
we have
\begin{align*}
\sum_{k = 1}^{N + k} \frac{|c_k|^2}{\tilde{p}_k}  %
&= \left( \sum_{k = 1}^{N + k} \frac{|c_k|^2}{\tilde{p}_k} \right) \left( \sum_{k = 1}^{K + N} \tilde{p}_k \right) \\ %
&\geq \left( \sum_{k = 1}^{N + k} \frac{|c_k|}{\sqrt{\tilde{p}_k}} \sqrt{\tilde{p}_k} \right)^2 \\ %
&= \left( \sum_{k = 1}^{N + k} |c_k| \right)^2,
\end{align*}
where the lower bound is attained if and only if there is an
$\alpha \in \mathbb{C}$ such that
\begin{align*}
\frac{|c_k|^2}{\tilde{p}_k} = \alpha \tilde{p}_k, \quad k = 1, \dots, K + N, %
\end{align*}
or
\begin{align*}
\tilde{p}_k = \frac{|c_k|}{\sqrt{\alpha}}, \quad k = 1, \dots, K + N. %
\end{align*}
The proportionality constant $\alpha$ can be determined from the
condition $\sum\nolimits_{k = 1}^{K + N} \tilde{p}_k = 1$, giving
\begin{align} \label{eq:8}
\tilde{p}_k = \frac{|c_k|}{\sum_{i = 1}^{K + N} |c_i|}, \quad k =
1, \dots, K + N.
\end{align}
Putting this expression in \eqref{eq:7} gives the reduced problem
\begin{align*}
\begin{array}{cl}
\min\limits_{c_1, \dots, c_{K + N}} & \left( \sum\limits_{k = 1}^{K + N} |c_k| \right)^2 \\
\text{s.t.}                         & \bm{\tilde{A}}^H \bm{c} = \bm{y}, %
\end{array}
\end{align*}
or, equivalently,
\begin{align} \label{eq:9}
\begin{array}{cl}
\min\limits_{c_1, \dots, c_{K + N}} & \sum\limits_{k = 1}^{K + N} |c_k|  \\
\text{s.t.}                         & \bm{\tilde{A}}^H \bm{c} = \bm{y}. %
\end{array}
\end{align}
This is a complex-valued $l_1$-optimization problem, hence it can
be expected to give a sparse solution in $\{c_k\}$. This, in turn,
gives a sparse solution in $\{\tilde{p}_k\}$ through \eqref{eq:8},
and thus in
\begin{align*}
p_k = \frac{\tilde{p}_k}{w_k} = \frac{|c_k| \|\bm{y}\|_2^2}{\|\bm{a}_k\|_2^2 \sum_{i = 1}^{K + N} |c_i|}, \quad k = 1, \dots, K + N. %
\end{align*}
%

To explore the behavior of SPICE in more detail, we can notice, by
denoting first $K$ components of the $k$-th row of
$\bm{\tilde{A}}^H$ as $\bm{\varphi}_k^H$, i.e., $\bm{\varphi}_k^H
:= [(\bm{\tilde{a}}_1)_k \; \cdots \; (\bm{\tilde{a}}_K)_k]$, and observing that
the constraints in \eqref{eq:9} read $c_{k+j} = y_j - \bm{\varphi}_j^H \bm{\tilde{c}}$
for $j = 1, \dots, N$, that \eqref{eq:9} is equivalent to
\begin{align*}
\min\limits_{c_1, \dots, c_K} \quad \sum\limits_{k = 1}^N |y_k - \bm{\varphi}_k^H \bm{\tilde{c}}| + \sum\limits_{k = 1}^K |c_k|, %
\end{align*}
where $\bm{\tilde{c}} := [c_1 \; \cdots \; c_K]^T$, or more compactly
\begin{align} \label{eq:11}
\min\limits_{\bm{\tilde{c}}} \quad \|\bm{y} - \bm{\Phi} \bm{\tilde{c}}\|_1 + \|\bm{\tilde{c}}\|_1, %
\end{align}
where $\bm{\Phi}^H := [\bm{\varphi}_1 \; \cdots \;
\bm{\varphi}_N]$, i.e., $\bm{\Phi}$ corresponds to the first $K$
columns of $\bm{\tilde{A}}^H$. Equation~\eqref{eq:11} is
essentially a simplified (complex-valued) version of the
LAD-Lasso~\citep{Wang-Li-Jiang-07} or the
RLAD~\citep{Wang-Gordon-Zhu-06}, where $\bm{\tilde{c}}$ takes the
role of a parameter vector, and the regressors have been scaled by
$w_k^{-1/2} = \|\bm{y}\|_2 / \|\bm{a}_k\|_2$, so that their
Euclidean norms are equal to $\|\bm{y}\|_2$. The fact that the
cost function in \eqref{eq:11} considers the $\ell_1$ norm of the
residuals ($\bm{y} - \bm{\Phi} \bm{\tilde{c}}$) instead of their
$\ell_2$ norm suggests that SPICE might be a robust estimator
against outliers or errors with heavy-tailed distributions (since,
heuristically speaking, it does not penalize large deviations of
the residuals from zero, due mainly to outliers, as much as the
$\ell_2$ norm); in fact, this is the reason why some authors have
proposed the use of the LAD-Lasso instead of the standard Lasso in
the presence of outliers~\citep{Wang-Li-Jiang-07}.


We can summarize these results in the following theorem:

\begin{thm} \label{thm:1}
The limit value of the SPICE iterations (allowing for different
$\sigma_k$), which corresponds to the minimizer of \eqref{eq:2},
is also given by the minimizer of \eqref{eq:11}, by performing the
following change of variables:
\begin{multline*}
p_k = \frac{\|\bm{y}\|_2^2 |c_k|}{\|\bm{a}_k\|_2^2 \left\{\sum_{i = 1}^K |c_i| + \sum_{k = 1}^N |y_k - \bm{\varphi}_k^H \bm{\tilde{c}}| \right\}}, \\ %
k = 1, \dots, K + N. %
\end{multline*}
\end{thm}

\subsection{Equal variances} \label{subsec:equal_sigma}
Now we will analyze the variant of SPICE where the variances are
constrained to be equal. The development in this case is exactly
as in Section~\ref{subsec:different_sigma} until
equation~\eqref{eq:7}. At this point, the constraint $\sigma_1 =
\cdots = \sigma_N =: \sigma$ implies that $\tilde{p}_{K+1} =
\cdots = \tilde{p}_{K+N}$, which allows us to simplify
\eqref{eq:7} as
\begin{align} \label{eq:7b}
\begin{array}{cl}
\min\limits_{\begin{array}{c} p'_1, \dots, p'_{K + 1} \geq 0, \\ c_1, \dots, c_{K + N} \end{array}} & \displaystyle \sum\limits_{k = 1}^K \frac{\displaystyle |c_k|^2}{\displaystyle p'_k} + \frac{\displaystyle N}{\displaystyle p'_{K+1}} \sum_{k=K+1}^{K+N} |c_k|^2 \\ %
\text{s.t.}                                                                                         & \sum\limits_{k = 1}^{K + 1} p'_k = 1 \\ %
                                                                                                    & \bm{\tilde{A}}^H \bm{c} = \bm{y}, %
\end{array}
\end{align}
where $p'_k = \tilde{p}_k$ for $k = 1, \dots, K$, $p'_{K+1} = N
\tilde{p}_{K + 1}$, and $\bm{c} := [c_1 \; \cdots \; c_{K +
N}]^T$. Now, the Cauchy-Schwarz argument used in
Section~\ref{subsec:different_sigma} reveals that
\begin{align*}
p'_k = \left\{ \begin{array}{ll}
\displaystyle \frac{|c_k|}{\sqrt{\alpha}}, \quad & k = 1, \dots, K, \\ %
\displaystyle \sqrt{\frac{N}{\alpha} \sum_{k=K+1}^{K+N} |c_k|^2}, \quad & k = K + 1, %
\end{array} \right.
\end{align*}
and from the condition $\sum\nolimits_{k = 1}^{K + N} \tilde{p}_k
= 1$ we obtain
\begin{align} \label{eq:8b}
\alpha = \left( \sum_{k = 1}^K |c_k| + \sqrt{N \sum_{k=K+1}^{K+N} |c_k|^2} \right)^2. %
\end{align}
The constants $c_k$, on the other hand, must be the solution of
\begin{align} \label{eq:9b}
\begin{array}{cl}
\min\limits_{c_1, \dots, c_{K + N}} & \sum\limits_{k = 1}^K |c_k| + \sqrt{N \displaystyle \sum_{k=K+1}^{K+N} |c_k|^2}  \\
\text{s.t.} & \bm{\tilde{A}}^H \bm{c} = \bm{y}. %
\end{array}
\end{align}
Just as in Section~\ref{subsec:different_sigma}, \eqref{eq:9b} can
be rewritten as
\begin{align*}
\min\limits_{\bm{\tilde{c}}} \quad \sqrt{N \displaystyle \sum_{k=1}^{N} |y_k - \bm{\varphi}_k^H \bm{\tilde{c}}|^2} + \|\bm{\tilde{c}}\|_1, %
\end{align*}
where $\bm{\tilde{c}} := [c_1 \; \cdots \; c_K]^T$, or
\begin{align} \label{eq:11b}
\min\limits_{\bm{\tilde{c}}} \quad \sqrt{N} \|\bm{y} - \bm{\Phi} \bm{\tilde{c}}\|_2 + \|\bm{\tilde{c}}\|_1. %
\end{align}
Equation~\eqref{eq:11b} is essentially a simplified
(complex-valued) version of the standard
Lasso~\cite{Tibshirani-96}, where $\bm{\tilde{c}}$ takes the role
of a parameter vector, and the Euclidean norms of the regressors
have been equalized. We summarize these results as
a theorem:

\begin{thm} \label{thm:2}
The limit value of the SPICE iterations (imposing the constraint
that $\sigma_1 = \cdots = \sigma_N$), which corresponds to the
minimizer of \eqref{eq:2}, is also given by the minimizer of
\eqref{eq:11b}, by performing the following change of variables:
\begin{align*}
p_k     &= \frac{\|\bm{y}\|_2^2 |c_k|}{\|\bm{a}_k\|_2^2 (\|\bm{\tilde{c}}\|_1 + \sqrt{N} \|\bm{y} - \bm{\Phi} \bm{\tilde{c}}\|_2)}, \quad k = 1, \dots, K \\ %
p_{K+1} &= \frac{\sqrt{N} \|\bm{y}\|_2^2 \|\bm{y} - \bm{\Phi} \bm{\tilde{c}}\|_2}{N (\|\bm{\tilde{c}}\|_1 + \sqrt{N} \|\bm{y} - \bm{\Phi} \bm{\tilde{c}}\|_2)}. %
\end{align*}
\end{thm}

The following remarks are appropriate:

\begin{remm}
The results stated in Theorems~\ref{thm:1} and \ref{thm:2} are quite
surprising, because they reveal that different assumptions on the
noise variance produce versions of SPICE which are equivalent to
two quite different but standard sparse estimators, namely the
LAD-Lasso and the Lasso.
\end{remm}

\begin{remm}
Even though the equivalent Lasso formulations are not given in the same
variables as the SPICE method, the required variables
transformations (between the $c_k$'s and the $p_k$'s) are simple
scalings. This means that the sparsity properties of SPICE are
essentially the same as the ones for the equivalent Lasso
estimators.
\end{remm}

\begin{remm}
The relations between the $c_k$'s and the $p_k$'s given by Theorems~\ref{thm:1}
and \ref{thm:2} have a nontrivial structure, which comes from the fact that
SPICE considers the (unknown) noise variances as parameters to be estimated,
and puts them in the same footing as the amplitudes of the spectral lines.
\end{remm}

\begin{remm}
The cost function $g(\bm{R})$ minimized by SPICE in \eqref{eq:g(R)}
can be interpreted as follows: The first term of $g(\bm{R})$, 
$\bm{y}^H \bm{R}^{-1} \bm{y}$, is a model fit measure, while the second
term, $\|\bm{y}\|_2^{-2} \tr \bm{R}$, can be interpreted as a trace
heuristic or nuclear norm regularization (since $\bm{R} = \bm{R}^H \geq 0$,
so the trace and nuclear norm coincide) \citep{Fazel-Hindi-Boyd-01}.
This regularization term is known to encourage low rank matrices $\bm{R}$,
which, due to its structure, $\bm{R} = \bm{A}^H \bm{P} \bm{A}$, enforces
the vector $[p_1, \dots, p_{K+N}]^T$ to be sparse. This interpretation thus
provides an alternative heuristic justification for the sparsity-inducing
behavior of SPICE.
\end{remm}

\begin{remm}
Theorems~\ref{thm:1} and \ref{thm:2} have been presented for the complex-valued
versions of SPICE. However, the derivations in this section apply almost unaltered
to real valued problems. This means that Theorems~\ref{thm:1}
and \ref{thm:2} establish Lasso-type equivalences for the real-valued versions
of SPICE as well. Notice, however, that the complex Lasso versions
can be seen as real-valued Group Lasso estimators, as explained next.
\end{remm}

\begin{remm}
The complex-valued nature of SPICE is inherited by its Lasso equivalents.
Thus, for example problem~\eqref{eq:11b} does not behave as the standard
(real-valued) Lasso, but as the (real-valued) Group Lasso~\citep{Yuan-Lin-06}.
To see this, let us define
\begin{align*}
\bm{y}_R &:= \left[ \begin{array}{c} \Real \bm{y} \\ \Imag \bm{y} \end{array} \right], %
\qquad \bm{\tilde{c}}_R := \left[ \begin{array}{c} \Real \bm{\tilde{c}} \\ \Imag \bm{\tilde{c}} \end{array} \right] \\ %
\bm{\Phi}_R &:= \left[ \begin{array}{cc} %
\Real \bm{\Phi} & -\Imag \bm{\Phi} \\
\Imag \bm{\Phi} & \Real \bm{\Phi}
\end{array} \right]
\end{align*}
Based on this notation, \eqref{eq:11b} can be written as
\begin{align} \label{eq:real_11b}
\min\limits_{\bm{\tilde{c}}_R} \quad \sqrt{N} \|\bm{y}_R - \bm{\Phi}_R \bm{\tilde{c}}_R\|_2 + %
\sum_{k = 1}^K \left\| \left[ %
\begin{array}{c} %
(\bm{\tilde{c}}_R)_k \\
(\bm{\tilde{c}}_R)_{k + K}
\end{array} %
\right] \right\|_2. %
\end{align}
The second term in \eqref{eq:real_11b} is a sum of Euclidean
norms, which promotes group sparsity, i.e., it tries to enforce
that both the real and imaginary parts of individual entries of
$\tilde{c}$ become zero simultaneously. Similarly, \eqref{eq:11}
corresponds to a grouped version of the LAD-Lasso.
\end{remm}

\begin{remm}
Recently, a re-weighted version of SPICE, called LIKES, has been
proposed in \citep{Stoica-Babu-12}. We will not address here
the relation between LIKES and standard sparse estimators (such as
Sparse Bayesian Learning (SBL) and Automatic Relevance
Determination (ARD) \citep{Wipf-Nagarajan-08}), because this has partly
been discussed in \citep{Stoica-Babu-12}, and the equivalence to Lasso-type
estimators can be formally studied along the lines of \citep{Wipf-Nagarajan-08}.
\end{remm}

%
%

\section{Simulation Example} \label{sec:example}
In this section, a numerical example, based on \citep[Section IV]{Stoica-Babu-Li-11},
is used to illustrate the equivalence between SPICE and the LAD-Lasso,
formally established in Theorem~\ref{thm:1}.

Let $\bm{y}_k = y(t_k)$, $k= 1, \dots, N$, be the $k$-th sample, where the $t_k$'s
are irregular time samples, drawn independently from a uniform distribution on $[0, 200]$.
The basis functions considered here are of the form
\begin{align*}
\bm{a}_k = [e^{j \omega_k t_1} \; \cdots \; e^{j \omega_k t_N}]^T,
\end{align*}
where $\omega_k := 2 \pi k / 1000$. Following \citep{Stoica-Babu-Li-11}, we take $N = 100$,
and $\bm{y}$ to be given by \eqref{eq:model} with $K = 3$, $s_{145} = 3 e^{j \phi_1}$, $s_{310}
= 10 e^{j \phi_2}$ and  $s_{315} = 10 e^{j \phi_3}$, and $s_k = 0$ otherwise. The phases
$\phi_1$, $\phi_2$ and $\phi_3$ are independent random variables, uniformly distributed
in $[0, 2\pi]$. The noise $\epsilon$ is assumed to have a covariance matrix $0.25 \bm{I}$.

\begin{figure}
\centering
\includegraphics[height=0.75\columnwidth]{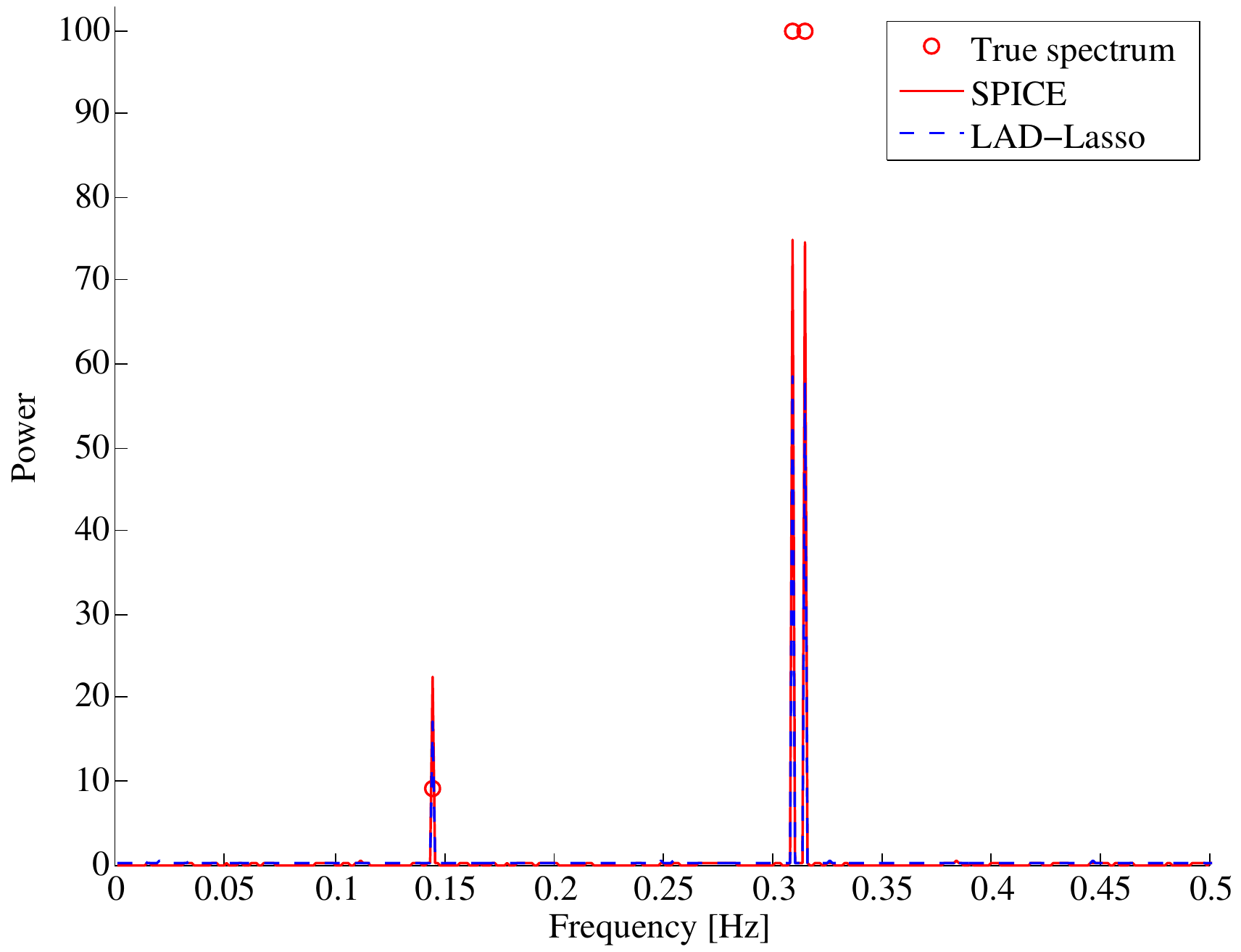}
\caption{Spectrum obtained by SPICE and LAD-Lasso.}
\label{fig:1}
\end{figure}

The results of applying $100$ iterations of SPICE, \eqref{eq:SPICE}, and its LAD-Lasso equivalent \eqref{eq:11}, solved using the CVX package~\citep{Grant-Boyd-08}, are presented in Figure~\ref{fig:1}. As the figure shows, both estimators practically coincide, their differences being mainly due to numerical implementations. Notice also that these estimators correctly detect the location of the peaks of the true spectrum, even though the estimated amplitudes do not approach their true values; this observation is consistent with theoretical results regarding the bias of the Lasso and its variants~\citep{Buhlmann-vandeGeer-11}. On a PC with an 2.53 GHz Intel Core Duo CPU and 4 Gb RAM, $100$ iterations of SPICE take $23.0$ s, while the implementation of LAD-Lasso using CVX only takes $14.6$ s. However, if $N$ is further increased to $1000$, CVX is incapable of solving the LAD-Lasso problem, while SPICE can still provide a good (and numerically reliable) estimate.

\section{Conclusion} \label{sec:conclusions}
In this manuscript, the recently proposed SPICE method for sparse
estimation has been studied, and its relation to Lasso-type estimators
has been established. This connection may enable the use of existing
theoretical results for the Lasso to predict the behavior of SPICE in diverse
problem settings, and, at the same time, the application of the computationally
efficient algorithm developed for SPICE to sparse estimation problems where
the Lasso algorithms are currently impractical.

As a interesting future line of research, the relation between SPICE and
the Group Lasso suggests that the former method could be modified to deal
with general group sparsity problems (instead of only groups with two real
variables). In addition, from this relation it is easy to modify SPICE in order
to compensate for deficiencies already detected in standard Lasso estimators, 
such as lack of consistency in sparse support recovery, which can be fixed
by adding re-weighting steps (see, e.g., \citep{Zou-06}).

\appendix[Proof of Equation (5)] \label{app:Gauss-Markov}
In this Appendix we prove \eqref{eq:5}. Without loss of generality
we can assume that the values of $k$ for which $\tilde{p}_k = 0$
have been removed from the sum. We start by rewriting \eqref{eq:5}
as
\begin{align} \label{eq:app_1} %
\bm{y}^H (\bm{\tilde{A}}^H \bm{\tilde{P}} \bm{\tilde{A}})^{-1} \bm{y} %
= \min\limits_{\bm{c}} \bm{c}^H \bm{\tilde{P}}^{-1} \bm{c} \quad \text{s.t.} \quad \bm{\tilde{A}}^H \bm{c} = \bm{y}, %
\end{align}
where $\bm{\tilde{P}} := \diag(\tilde{p}_1, \dots, \tilde{p}_{K +
N})$. We will proceed by establishing the minimum value of the
right hand side of \eqref{eq:app_1} and showing that it coincides
with its left hand side. To this end, notice that since that
optimization problem is convex, $\bm{c}$ is an optimal solution of
the right hand side of \eqref{eq:app_1} if and only if there is a
Lagrange multiplier $\bm{\lambda} \in \mathbb{C}^N$ such that
\begin{align} \label{eq:app_2}
\frac{\partial}{\partial \bm{c}} [\bm{c}^H \bm{\tilde{P}}^{-1} \bm{c} + \bm{\lambda}^H (\bm{\tilde{A}}^H \bm{c} - \bm{y})] = 0, \qquad %
\bm{\tilde{A}}^H \bm{c} = \bm{y}, %
\end{align}
or, equivalently,
\begin{align} \label{eq:app_3}
2 \bm{\tilde{P}}^{-1} \bm{c} + \bm{\tilde{A}} \bm{\lambda} = 0, \qquad %
\bm{\tilde{A}}^H \bm{c} = \bm{y}. %
\end{align}
From this set of equations we obtain
\begin{align} \label{eq:app_4}
\bm{\lambda} &= - 2 (\bm{\tilde{A}}^H \bm{\tilde{P}} \bm{\tilde{A}})^{-1} \bm{y} \\
\bm{c} & = \bm{\tilde{P}} \bm{\tilde{A}} (\bm{\tilde{A}}^H \bm{\tilde{P}} \bm{\tilde{A}})^{-1} \bm{y}, \nonumber %
\end{align}
and the optimal cost of right hand side of \eqref{eq:app_1} gives
$\bm{c}^H \bm{\tilde{P}}^{-1} \bm{c} = \bm{y}^H (\bm{\tilde{A}}^H
\bm{\tilde{P}} \bm{\tilde{A}})^{-1} \bm{\tilde{A}}^H
\bm{\tilde{P}} \bm{\tilde{P}}^{-1} \bm{\tilde{P}} \bm{\tilde{A}}
(\bm{\tilde{A}}^H \bm{\tilde{P}} \bm{\tilde{A}})^{-1} \bm{y} =
\bm{y}^H (\bm{\tilde{A}}^H \bm{\tilde{P}} \bm{\tilde{A}})^{-1}
\bm{y}$, which corresponds to the left hand side of
\eqref{eq:app_1}. This concludes the proof of \eqref{eq:5}.

\begin{remm}
Equation~\eqref{eq:5} is closely related to the
so-called Gauss-Markov theorem, which states that, in a linear
regression framework, the least squares estimator is the minimum
variance unbiased estimator \citep{Kay-93}. In fact, let $\bm{z} =
\bm{\tilde{A}} \bm{\theta} + \bm{e}$, where $\bm{\theta} \in
\mathbb{C}^{K + N}$, $\bm{e} \sim \mathcal{CN}(\bm{0},
\bm{\tilde{P}}^{-1})$. Furthermore, suppose we are interested in
estimating $x = \bm{y}^H \bm{\theta}$. Then, the cost function in
the right hand side of \eqref{eq:5} can be interpreted as the
variance of an estimate $\hat{x} = \bm{c}^H \bm{z}$ of $x$, and
the corresponding constraint $\bm{\tilde{A}}^H \bm{c} = \bm{y}$
restricts $\hat{x}$ to be unbiased, while the left hand side of
\eqref{eq:5} corresponds to the minimum achievable variance,
according to the Gauss-Markov theorem.
\end{remm}


\bibliographystyle{IEEEtran}
\bibliography{cristian}
\end{document}